# Video Analysis of "*YouTube Funnies*" to Aid the Study of Human Gait and Falls – Preliminary Results and Proof of Concept

Babak Taati [1,2], Pranay Lohia [2,3], Avril Mansfield [1,2,4], Ahmed B. Ashraf [1]
[1] Toronto Rehabilitation Institute, [2] University of Toronto,
[3] Indian Institute of Technology, Kharagpur, [4] Sunnybrook Research Institute

*Abstract*— Because falls are funny, YouTube and other video sharing sites contain a large repository of real-life falls. We propose extracting gait and balance information from these videos to help us better understand some of the factors that contribute to falls. Proof-of-concept is explored in a single video containing multiple (n=14) falls/non-falls in the presence of an unexpected obstacle. The analysis explores: computing spatiotemporal parameters of gait in a video captured from an arbitrary viewpoint; the relationship between parameters of gait from the last few steps before the obstacle and falling vs. not falling; and the predictive capacity of a multivariate model in predicting a fall in the presence of an unexpected obstacle. Homography transformations correct the perspective projection distortion and allow for the consistent tracking of gait parameters as an individual walks in an arbitrary direction in the scene. A synthetic top view allows for computing the average stride length and a synthetic side view allows for measuring up and down motions of the head. In leave-one-out cross-validation, we were able to correctly predict whether a person would fall or not in 11 out of the 14 cases (78.6%), just by looking at the average stride length and the range of vertical head motion during the 1-4 most recent steps prior to reaching the obstacle.

## I. Introduction

Falling can be a serious health and safety issue, particularly for individuals with impaired balance control, such as older adults or people with chronic conditions affecting gait control. Studying human balance control and mechanisms leading to falls can help develop technologies and interventions for fall prevention [1]. However, falls are relatively rare and unpredictable events. Therefore, studying falls relies on either: 1) data collection over long periods of time with large sample sizes so as to capture sufficient numbers of falls; or 2) inducing falls in the laboratory with controlled postural perturbations.

In a prominent example of the first approach, researchers visually analyzed 227 falls recorded by security cameras at two nursing homes over 3+ years [2]. This approach is challenging due to the amount time required to match surveillance footage with nursing staff log files in order to find and analyze interesting events. The alternative approach of laboratory testing avoids this pitfall. Within the lab, balance perturbations are provided by devices such as moveable platforms or cable and pulley systems [3]. Various tools (e.g., motion capture, force plates) are used to capture kinematic, kinetic, and electrophysiological features in detail. However, investigators often receive criticism from clinicians regarding this approach; the concern is that postural perturbations do not adequately mimic in real-life falls.

It is, therefore, worth exploring methods to enable large-scale quantitative analysis of real-life falls outside the laboratory. The challenge in doing so is twofold: having access to a large number of real-life falls, and developing and validating computer-based methods to expedite the processing of such videos and reduce the burden of manual coding.

While falls have serious consequences, many people perceive falls as funny, especially when they do not result in an injury. Thousands of videos featuring falls in a variety of situations are uploaded to YouTube and other video sharing websites every year. Recent advances in computer vision algorithms facilitate automated, or semi-automated, processing of videos of human gait in natural settings [4-6]. We, therefore, propose using computer vision techniques to analyze real-life falls to support the ecological validity of laboratory findings.

The way people walk is influenced by attention, age, and various nervous or musculoskeletal conditions (e.g., hemiparetic gait following a stroke), all of which influence the risk of falling when faced with an unexpected obstacle. Furthermore, it has previously been shown that computer-based visual analysis can reveal factors such as age [7] and impaired gait [8]. We, therefore, hypothesize that the immediate spatiotemporal parameters of gait, i.e., those extracted from analysis of the last few strides before an unexpected obstacle, would be a mild predictor of whether the person would fall or not. This paper analyzes a single video and lacks statistical power to test the hypothesis; however, it serves as an important proof-of-concept and provides preliminary results in line with our hypothesis. To our knowledge, the analysis of YouTube funnies to study gait and falls is novel and has not been explored before.

## II. Method

### A. Video Selection

A single video with multiple falls was selected for this proof-of-concept study. When searching the internet to select this video, the initial criteria to select a suitable video were: 1) having a stable view point; 2) showing multiple people walking over the same falling hazard, e.g. a patch of ice or a step, where some people fall and others do not; and 3) showing a few steps of each person before they reach the falling hazard.

Stabilizing the view is possible in post-processing but requires additional processing. Stable views (e.g. security camera footages) were therefore preferred to handheld videos. The second condition allows for the comparison of fallers vs. non-fallers, and the third condition enables the processing of gait prior to the fall. YouTube and Dailymotion were searched with keywords such as falls/falling, tripping, slipping, etc.

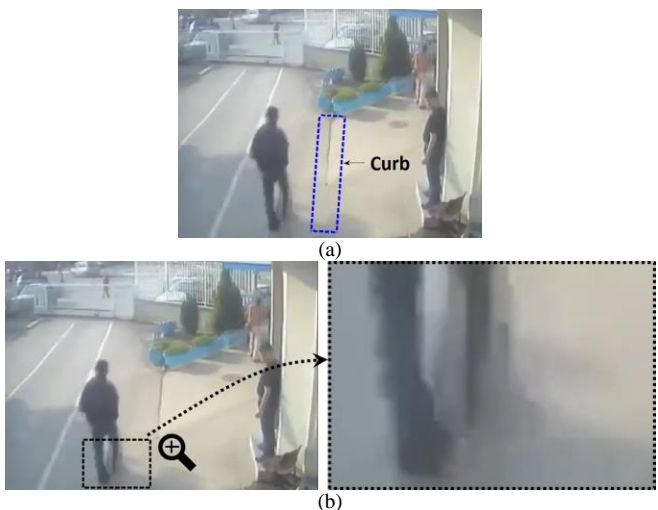

Figure 1. (a) A sample image from the video (frame # 1,882), showing a person walking from left to right towards the falling hazard, and a magnified view of the person's feet; (b) the curb is a falling hazard (highlighted).

We selected a single video that met our criteria [9]. The video is 3 minutes and 16 seconds long (4,370 frames at 25 frames per second). The obstacle in this video is a curb and is highlighted in Figure 1-a. People walking from the left side to the right face this falling hazard as a step up and people walking from the right side to the left will face an unexpected step-down. While the video is posted on YouTube with 480p resolution, the actual image quality is very poor and highly degraded. Figure 1-b illustrates the high level of spatial and temporal blurring due to compression. The video shows several pedestrians walking, as well as others sitting or biking. Of these, 16 people were identified as individuals crossing the obstacle while walking. For two people, less than a single stride was visible before reaching the obstacle; so only the remaining 14 were included in the analysis. Each person was assigned an identification (ID) number based on their order of appearance in the video. Table I presents relevant information about these 14 walking sequences.

### B. Head and Feet Tracking

The positions of the head and the two feet were manually marked in each frame. We emphasize that future work with higher resolution videos will allow for the automated detection and tracking of these and other body parts using computer-vision algorithms. Frames at which each foot was in contact with the ground were also manually marked.

### C. Spatiotemporal Parameters of Gait

Spatiotemporal parameters of gait that are potentially useful are the step length, step width, body sway, step symmetry, etc. We note, however, that our data is not amenable to computing some of these parameters. Reliably calculating the body sway, for instance, is difficult from a side view. We also note that the video is captured from an unknown angle and distance from the scene, and estimating absolute distances (e.g., in metres) is not possible. However, for comparison between pedestrians, relative distances are sufficient and we can, therefore, measure distances in pixels or relative to a person's height.

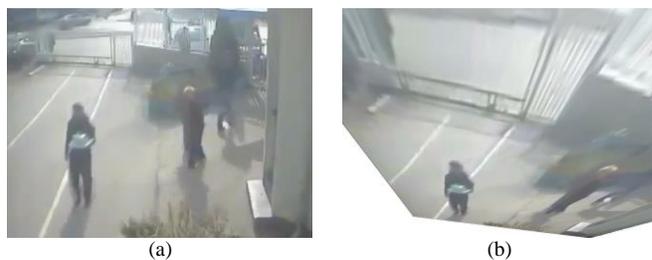

Figure 2. (a) A sample frame showing a person walking from right to left towards the walking hazard; (b) top-view homography of the same image frame (cropped).

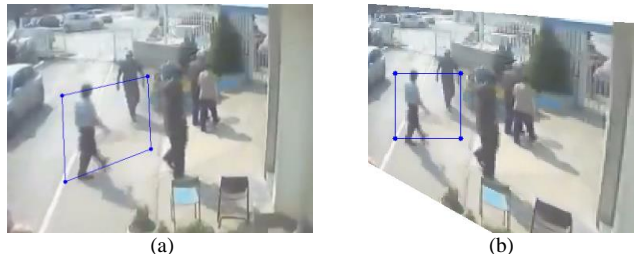

Figure 3. (a) A sample frame showing a person walking from left to right towards the walking hazard; (b) side-view homography of the same image frame (cropped).

Raw pixel coordinates could be misleading because of perspective projection distortion. The same pixel distance in image coordinates corresponds to larger physical distance for items or individuals closer to the camera than for those further away. Step lengths, head motions, and other measures are therefore affected as a person walks closer or further away from the camera, and direct computation of these features will result in inconsistent parameters of gait, highly dependent on where and in which direction a person walks.

We, therefore, employ a computer vision technique known as homography to transform the image frames such that computed gait parameters are stable and unaffected by perspective distortion. Using a homography to measure spatiotemporal parameters of gait has been previously validated [10]. We calculate two homographies: one to transform the images as if they were taken from the top view, and another to transform the images as if they were taken from the side view perpendicular to the walking direction.

The two visible traffic lanes visible on the left side of the scene were used to calculate the top-view homography. The two lines should appear in parallel in the transformed view. The same homography transformation was valid for the entire length of the video. This homography correctly maps image points on the ground plane (i.e. surface of the street) but distorts other points. It is, therefore, useful to compute features from only those points at which the person's foot was in contact with the ground. Figure 2 illustrates a sample image from the video (frame # 21) and its top-view transformation.

The side-view homography is person dependent. For each person, a quadrilateral was formed by the pixel coordinates of their head and their feet (average of left and right foot) at the start and at the end frame of their walk. The homography was calculated to map this quadrilateral into a rectangle. In this view, the vertical motions of the feet and the head are valid and unaffected by perspective projection, but other points in the image are distorted by the transformation. The horizontal displacement of head and feet are also incorrect because the width of the target rectangle is arbitrary. Figure 3 illustrates a sample image from the video and its side-view transformation.

Two features were extracted from the walking sequence of each person prior to the obstacle. The average stride-length (in pixels) was calculated using the top-view. The range of head motion (as a percentage of a person's height) was calculated using the side view. Gaussian smoothing was applied to manually annotated head coordinates prior to computing the range of head motions. The smoothing was slight ($\sigma=2$) and only removed the jitter from the manual annotation. To smooth the small jittering in the manual annotation of the feet, a median filter was applied to the pixel coordinates of each foot stance. Table 2 presents the value of these two features for each of the 14 people in the video.

It is possible to conceive of other spatiotemporal gait parameters that could potentially be calculated from the top and side views. From the top view, for instance, the average step length (along the direction of walking), and average step width (perpendicular to the direction of walking) can be calculated. These, however, would require an accurate estimate of the direction of walking, which is not necessarily possible when only a few steps are visible prior to the obstacle. Similarly, the average walking speed (pixels/second) would also require an accurate estimate of the walking direction. By contrast, the average stride length, the feature used in this study, measures the average distance between footfall locations and does not require an accurate estimate of the walking direction.

*D. Analysis*

Two standard binary classification methods were examined in this study: Support Vector Machines (SVM) and k-Nearest Neighbors (kNN) [11]. SVM was used with a linear kernel and with the default value of the regularization parameter (i.e., no specific tuning). kNN was used with the number of neighbors (k) equal to 3. Leave-one-out cross validation was used in both cases. That is, data from 13 people was used to train a classification model to predict fall/no-fall based on stride length and vertical head motion. This model was then tested on the left out 14th person to make a prediction. This procedure was repeated 14 times until performance was evaluated for all 14 people. This is standard evaluation methodology with limited data. The receiver operating characteristic (ROC) was generated based on both SVM and kNN outputs. An ROC curve illustrates the performance of a binary classifier, its trade-off between the True Positive Rate and the False Positive Rate when the classifier's discrimination threshold is varied. Area under the curve (AUC) values larger than 0.5 indicate better than chance performance and the maximum value of 1 indicates perfect performance. AUC values in the 0.7-0.8 range are typically interpreted as "fair" classification performance.

III. RESULTS

In leave-one-out cross-validation, SVM and k-NN were both able to correctly predict whether a person would fall or not in 11 out of the 14 cases (78.6%). The three people were misclassified in both classes were ID 1, ID 4, and ID 11. The AUC was 0.77 for SVM and 0.64 for kNN.

IV. DISCUSSION AND LIMITATIONS

In this proof of concept paper, we used a combination of manual annotation and standard image processing and computer vision methods to analyse a YouTube video of multiple falls. The novelty of this analysis is in exploring use of video data shared via social media platforms to study gait, balance and falling. We showed that informative spatiotemporal parameters of gait can be extracted even in a very low-resolution video. Using these two parameters, standard binary classification algorithms correctly predicted fall vs. no-fall in 11 out of 14 cases. While this is a preliminary analysis and lacks statistical power, results are in line with our initial hypothesis that immediate spatiotemporal parameters of gait would be a mild predictor of whether the person would fall or not in the presence of a falling hazard.

This analysis has some limitations. Of the 14 gait sequences included, 6 people carried something in their hands (or over their shoulder) when walking. These included a sweater or a jacket (ID 2 and 8), a bag or a purse (ID 3, 5, and 6), and a baby (ID 10). Carrying something, especially carrying a heavy load such as a baby can influence the way a person walks. Another person (ID 11) appears to slightly change their direction during their walk, which might influence their gait parameters. Finally, attention was not considered in this analysis. For example, two people (ID 13 and 14) might have noticed another person (ID 12) fall over the curb before they reached the falling hazard. This potentially drew their attention to the curb and might have influenced their gait. Nevertheless, these two people were correctly predicted to not fall based on their gait prior to reaching the obstacle, so knowledge of the tripping hazard might have influenced gait parameters.

Low image quality was not a major issue in this study and might indeed have been an inadvertent benefit. Computer vision human pose tracking algorithms work best in high resolution. So, in subsequent studies, when the goal is to analyse hundreds or thousands of videos automatically, it will be best to impose a minimum standard on image quality. For this proof-of-concept study, however, a fully automated method of video analysis was not essential since manual annotation was possible due to the relatively small number of image frames (<4,400). As an unexpected benefit, the very low resolution means that the people in the video are completely unidentifiable. For subsequent analysis of many high-resolution videos, only aggregate statistical results will be reported for privacy concerns. In this proof-of-concept study, however, we present detail analysis of a single video and individual falls and post our annotations and results online. It is, therefore, useful that our analysis does not risk any privacy concerns.

This is a small scale analysis using single video and 14 people walking over a falling hazard. As such, it is difficult to speculate whether results will generalize to a larger number of videos. Nevertheless, this study serves as a promising proof-of-concept to draw the attention of the research community to online falling videos.

Finally, we note that manual annotation (i.e. not automated computer vision-based methods) was used to mark the location of the feet and the head frame-by-frame. This will not be feasible when assessing a larger number of videos. Our intention is to use higher resolution videos and fully automated computer vision human pose detection and tracking. We also note that crowdsourcing (e.g. Mechanical Turk) is another possible option for inexpensively annotating a large number of videos in cases where automated tracking fails.

## V. CONCLUSIONS AND FUTURE WORK

In future work, we plan to use human pose detection and tracking algorithms to conduct a more automated analysis of a larger number of videos in the presence of various fall hazards: curbs, ice, snow, etc. We also plan to use natural language processing (NLP) techniques to browse through YouTube and other sites and automatically shortlist potentially appropriate videos for balance and gait analysis.

Future work will also expand to other applications that are difficult to study in laboratories, such as human motor development and mechanics of sports injuries. It is, for instance, difficult to convince small children to cooperate with motor control studies; but proud parents upload many videos showing their child's first steps. Similarly, large numbers of videos of real sports injuries are available on YouTube. Analyzing such videos could determine forces that cause bone fractures or concussions to help in designing protective equipment.

Finally, research in this area poses interesting computer vision challenges related to automated human pose detection and tracking. We hope this paper motivates further research into reliable and accurate pose tracking in low-resolution images and in videos with moving/shaking view from a handheld camera. This could advance the field of computer vision by posing challenging problems from a real life problem.

## ACKNOWLEDGMENT

The authors would like to thank Mitacs and the Toronto Rehabilitation Institute – University Health Network for funding support; equipment and space have been funded with grants from the Canada Foundation for Innovation, Ontario Innovation Trust, and the Ministry of Research and Innovation. Pranay Lohia's summer internship at the University of Toronto was supported by a Mitacs Globalink award. Avril Mansfield is supported by a New Investigator Award from the Canadian Institutes of Health Research (MSH-141983).

TABLE I. PERSONS WALKING OVER AN UNEXPECTED OBSTACLE

| Person ID | 1 | 2 | 3 | 4 | 5 | 6 | 7 | 8 | 9 | 10 | 11 | 12 | 13 | 14 |
|---|---|---|---|---|---|---|---|---|---|---|---|---|---|---|
| Start frame | 1 | 351 | 1,063 | 1,117 | 1,600 | 2,940 | 3,050 | 3,241 | 3,340 | 3,544 | 3,643 | 3,922 | 3,957 | 4,080 |
| End frame | 54 | 385 | 1,100 | 1,155 | 1,608 | 2,990 | 3,075 | 3,285 | 3,368 | 3,630 | 3,672 | 3,945 | 4,013 | 4,110 |
| Walking direction | ← | → | → | → | → | → | ← | → | → | → | → | → | → | ← |
| Fall / No-Fall | F | F | NF | F | F | NF | NF | NF | F | NF | NF | F | NF | NF |
| # Steps before the obstacle | 2 | 1 | 2 | 2 | 1 | 4 | 1 | 2 | 2 | 3 | 2 | 2 | 3 | 2 |
| # Frames before the obstacle | 54 | 35 | 38 | 39 | 9 | 51 | 26 | 45 | 29 | 87 | 30 | 24 | 57 | 31 |

TABLE II: AVERAGE STRIDE LENGTH (L) AND AVERAGE RANGE OF HEAD MOTION (H) FOR ALL 14 PERSONS

| Person ID | 1 | 2 | 3 | 4 | 5 | 6 | 7 | 8 | 9 | 10 | 11 | 12 | 13 | 14 |
|---|---|---|---|---|---|---|---|---|---|---|---|---|---|---|
| L (pixels) | 85.2 | 81.9 | 73.4 | 86.0 | 36.2 | 114.9 | 130.2 | 115.8 | 78.2 | 97.9 | 83.3 | 101.1 | 69.1 | 80.0 |
| H (% height) | 7.6 | 2.3 | 7.1 | 5.6 | 2.1 | 6.5 | 7.1 | 4.0 | 2.9 | 5.4 | 3.1 | 1.7 | 6.4 | 5.4 |